\pdfoutput=1
\documentclass{article}

\usepackage[preprint]{neurips_2026}

\usepackage[utf8]{inputenc}
\usepackage[T1]{fontenc}
\usepackage{hyperref}
\usepackage{url}
\usepackage{booktabs}
\usepackage{array}
\usepackage{amsfonts,amsmath,amssymb,amsthm}
\usepackage{nicefrac}
\usepackage{microtype}
\usepackage{xcolor}
\usepackage{graphicx}
\usepackage{caption}
\graphicspath{{figures/}}

\newtheorem{proposition}{Proposition}
\newtheorem{corollary}{Corollary}
\newcommand{\dprime}{$d'$}
\newcommand{\gpuhours}{51}

\title{Language Models Share a Readout Bottleneck\\in Working Memory}

\author{%
  Dongyu Gong \\
  Yale University \\
  \texttt{dongyu.gong@yale.edu} \\
}

\begin{document}

\maketitle

\begin{abstract}
Large language models (LLMs) reproduce a human working-memory signature on the N-back task: performance declines as memory load N increases, and errors concentrate on lures, items that repeat the current stimulus at a neighboring but incorrect serial position. We ask whether this shared behavioral signature reflects a shared mechanism. The literature offers two candidate accounts. On a \emph{precision} account, softmax attention cannot concentrate its weight on the item N positions back, so adjacent items are retrieved instead. On an \emph{interference} account, attention selects the correct item, but the content of competing items corrupts the decision. We dissociate these accounts in 15 pretrained LLMs from three families (70 million to 14 billion parameters) and in minimal transformers trained from scratch. In all three families, the attention heads that carry the task attend to the target on most missed matches. Linear probes decode near-perfectly whether a matching letter occurred at the target or at a lure position, even on trials where the model false-alarms to the lure, although its own decisions barely distinguish the two. Sharpening attention yields negligible gains, whereas suppressing the letter identity of non-target items improves sensitivity (\dprime{}) in nearly all model-by-load conditions. In larger models the task is spread over more heads, and removing interference helps more. In minimal transformers, precision limits a single attention layer, but a second layer removes this constraint. The working-memory limit of LLMs therefore lies in reading items out under interference rather than in locating them, consistent with interference accounts of human working memory.
\end{abstract}

\section{Introduction}
\label{sec:intro}

Working memory, the ability to hold a few pieces of information in mind and use them, is famously limited in humans \citep{cowan2001magical,ma2014changing}. A standard probe is the N-back task \citep{kirchner1958age}: people see a stream of letters and, for each one, report whether it matches the letter N positions earlier. Performance falls steeply as N grows, and false alarms occur most often on \emph{lures}, letters that repeat the current one at a nearby but wrong distance, such as N$-$1 or N$+$1 back \citep{kane2007nback}. A lure error shows that the letter was remembered but its position was confused, and such errors are usually explained as a failure to select the target against competing items \citep{engle2002executive,oberauer2012interference,szmalec2011interference}.

Large language models (LLMs) show the same signature \citep{gong2024chatgpt,hu2025nback,xiong2026superposition}. This is surprising, because an LLM does not need to hold earlier items in a limited store: the whole sequence stays in its input, and at every step it can look back at any earlier item through \emph{attention} \citep{vaswani2017attention}. Roughly, the current position gives every earlier position a score, the scores are turned into weights that sum to one, and the model takes the weighted average of what the earlier positions contain. Because nothing is forgotten, the limit must arise when the model retrieves an earlier item and uses it to decide.

The shared signature raises a question that is central to comparing neural systems, namely whether networks that behave alike also compute alike. Similar errors can arise from different mechanisms, and representations that look alike to a decoder need not play the same role in behavior. N-back is a useful test case, because two accounts that predict the same lure errors make opposite predictions about the computation inside the model, and these predictions can be checked across model families and sizes.

\begin{figure}[t]
  \centering
  \includegraphics[width=0.97\textwidth]{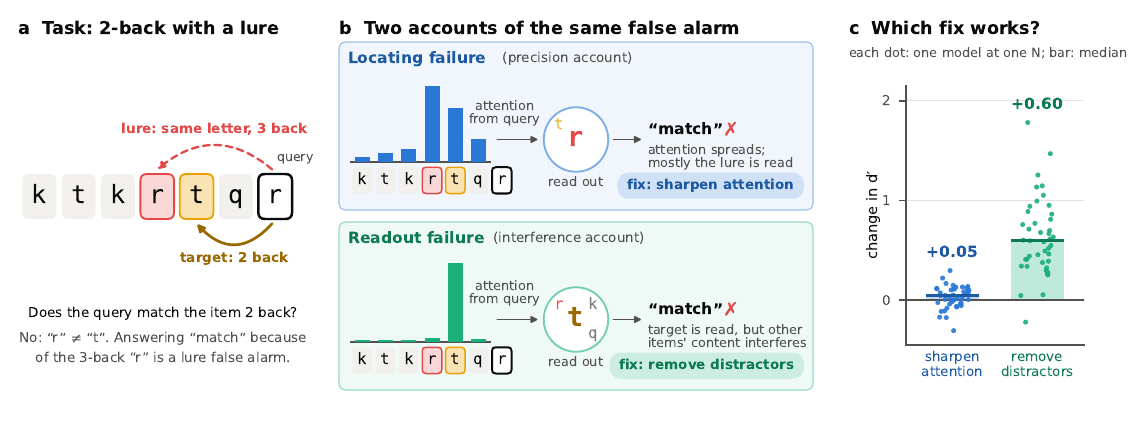}
  \caption{\textbf{Two accounts of the same error.} (a) In 2-back, the query ``r'' must be compared with the item two back (``t''). The ``r'' three back is a lure: answering ``match'' because of it is a lure false alarm. (b) Attention lands on the lure and reads its letter (precision account; fix: sharpen attention), or lands on the target while other items' letters leak into the decision (interference account; fix: remove the distractors' content). (c) Across 15 LLMs and N = 2--4 (dots: one model at one N; bars: medians), sharpening the task's heads barely changes sensitivity \dprime{}, whereas removing the letter identity of non-target items raises it in 42 of 43 cases (Section~\ref{sec:causal}).}
  \label{fig:overview}
\end{figure}

Two accounts of where this process fails have been proposed (Fig.~\ref{fig:overview}b). On the first, the \emph{precision} account, the model fails to locate the target. Attention weights are computed by a softmax, a form of divisive normalization \citep{carandini2012normalization}: each position's weight is its own (exponentiated) score divided by the sum over all positions, so positions compete for weight. The target's immediate neighbors lie only one position away from it (N$\pm$1 back), so if the model cannot tell adjacent positions apart sharply, they receive scores almost as high as the target's, and part of the weight spills onto them. A lure error then occurs because attention lands partly on the lure and the model reads its letter. This account was proposed for small transformers trained on N-back \citep{gong2024selfattention} and is supported by analyses of softmax \citep{velickovic2025softmax}. On the second, the \emph{interference} account, the model locates the target but fails to read it out correctly. Attention lands on the target, but inside the model the target's content is mixed with the content of other recent items, and the mixture corrupts the decision. \citet{xiong2026superposition} found evidence for this account: the models' errors are biased toward recent items, item representations overlap inside the models, and suppressing letter-identity directions can improve performance. But the precision account also predicts errors biased toward neighbors. Attention averaged over trials does peak at the target, but it also places some weight on the neighbors, and an average cannot show where attention goes on the trials where the model errs. The two accounts have not been tested against each other.

The two accounts make different predictions. If the precision account is right, attention should be off target when the model errs, and sharpening attention should help. If the interference account is right, attention should be on target even on errors, the information needed for the correct answer should be present in the model, and removing the competing content should help more than sharpening attention.

We test these predictions in 15 pretrained LLMs and in small transformers trained from scratch, where the precision account can be analyzed exactly. In every model we tested, the results favor the interference account. The attention heads that carry the task attend to the target on most errors, and a simple classifier reads from the model's own activations whether a repeated letter sat at the target or at a lure position, even on the trials where the model falls for the lure (Section~\ref{sec:locate}). Causally, erasing the content of competing items improves performance far more than sharpening attention does (Section~\ref{sec:causal}; Fig.~\ref{fig:overview}c). The same dissociation holds in every family and across a 200-fold range of size; with scale the task is spread over more heads, but the bottleneck stays the same. The precision account does apply to a transformer with a single attention layer, but a second layer removes this limit (Section~\ref{sec:theory}).

\section{Task, models and measures}
\label{sec:setup}

\textbf{The N-back task.} Each sequence has 24 letters drawn from 20 consonants. We call the current letter the \emph{query} and the letter N positions before it the \emph{target}. The correct answer is ``match'' if the query equals the target and ``no match'' otherwise. Each test sequence contains four lures: letters that repeat the query at N$\pm$1 or another non-target distance without creating a match.

\textbf{Presenting the task to LLMs.} The prompt begins with a one-sentence instruction and three solved example sequences, followed by the test sequence. Each line holds one letter and its answer, ``\texttt{m}'' for match or ``\texttt{-}'' for no match (for example, ``\texttt{t m}'' or ``\texttt{k -}''). The answers to earlier items are shown correctly, so the model never builds on its own mistakes. At each letter, we measure the model's preference for ``\texttt{m}'' over ``\texttt{-}'' as the difference between their log-probabilities. To turn this preference into a yes/no answer, we choose a threshold on 200 calibration sequences and apply it unchanged to 1,000 separate test sequences per model and N.

\textbf{Measuring performance.} As in human studies, we report sensitivity \dprime{} from signal detection theory \citep{hautus1995corrections}: how far apart the model's responses to matches and non-matches lie, in units of their spread. \dprime{} is 0 at chance and, unlike accuracy, is not inflated by a bias toward answering ``no match''. We also report hit rates and false-alarm rates, separately for lures and for other non-matches.

\textbf{Models.} We test 15 pretrained models from three families: Pythia, seven sizes from 70M to 6.9B parameters \citep{biderman2023pythia}; GPT-2, four sizes from 124M to 1.5B \citep{radford2019gpt2}; and Qwen2.5, four sizes from 0.5B to 14B \citep{qwen2025qwen25}. Section~\ref{sec:theory} also uses small transformers trained from scratch on N-back (Appendix~\ref{app:methods}).

\textbf{Attention heads and the residual stream.} A transformer is a stack of layers, each containing several \emph{attention heads}; each head computes its own attention pattern and writes its own output, and a few heads often do most of the work for a given task \citep{elhage2021framework,olsson2022induction,wu2024retrieval}. The model also keeps, for each position, a running vector that every layer reads from and adds its output to. This \emph{residual stream} is the model's working representation of that position. To find the heads that carry N-back, we use activation patching \citep{zhang2024patching}. We run the model on two sequences that differ only in the target letter, so that the correct answer flips, and copy the output of a single head from the first run into the second. A head whose output alone restores much of the first run's answer is important for the task. One head usually dominates and is the same across N; at N = 2 it alone recovers 16--73\% of the difference between the two runs, depending on the model.

\section{LLMs show a human-like load effect and lure errors}
\label{sec:behaviour}

\begin{figure}[t]
  \centering
  \includegraphics[width=0.88\textwidth]{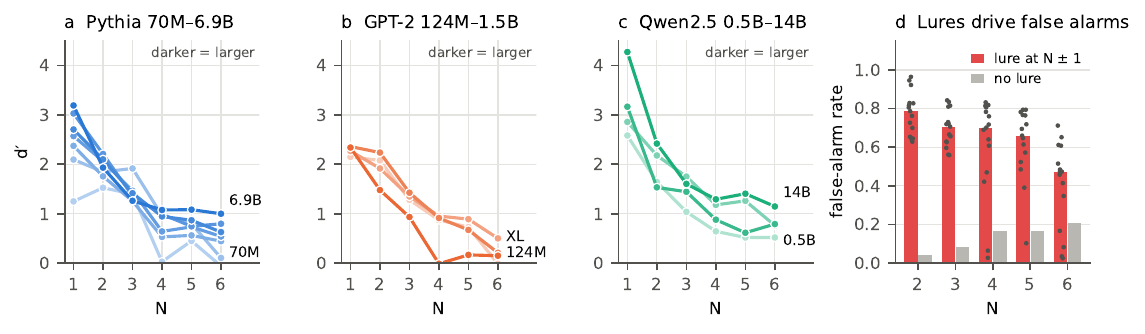}
  \caption{\textbf{LLMs decline with N and fall for lures.} (a--c) Sensitivity \dprime{} against N for three model families; darker lines are larger models. (d) False-alarm rates on lures at N$\pm$1 (red; dots are models) and on other non-matches (gray), medians over 15 models.}
  \label{fig:behaviour}
\end{figure}

As in earlier studies \citep{gong2024chatgpt,hu2025nback,xiong2026superposition}, all 15 models get worse as N grows (Fig.~\ref{fig:behaviour}a--c; Table~\ref{tab:l2}). Median \dprime{} falls from 2.57 at N = 1 (range 1.25--4.28) to 0.91 at N = 4 and 0.52 at N = 6. Scale helps when the task is easy but much less when it is hard: the largest model, Qwen2.5-14B, reaches 4.28 at N = 1 but only 1.15 at N = 6. As in humans and in \citet{xiong2026superposition}, errors concentrate on lures (Fig.~\ref{fig:behaviour}d). At N = 2, models answer ``match'' to a median 79\% of lures at N$\pm$1, against 4\% of other non-matches.

Two checks rule out trivial explanations. Weaker models can drift to an easier task \citep{hu2025nback}, but the $k$-back rule that best fits the answers differs from N in only 14 of 90 cases, all at N$\,\ge\,$4 in small models. And a chat format, one letter per turn as in \citet{gong2024chatgpt}, gives the same decline in Qwen2.5-Instruct 0.5B--7B (\dprime{} 1.8--2.0 at N = 1, 0.4--0.7 at N = 6).

These lure errors raise the question we address next: whether a model that false-alarms on a lure has attended to the wrong item, or has attended to the right item and misread it.

\section{LLMs find the target but misread it}
\label{sec:locate}

\textbf{Attention lands on the target, even on errors.} We first ask where the task's heads look. For each query, we measure how much attention the five most important heads send from the query letter to the target's line (its letter, answer and line break) and to the lines just before and after it. We say the heads \emph{locate} the target when its line receives at least as much attention as either neighbor.

On correctly answered matches, the heads locate the target on a median 88--91\% of trials (14 models; GPT-2 Large is excluded because its top heads attend to none of the three lines). On correctly answered non-matches, they do so on only about 25\%. The contrast indicates that the heads search by content, attending to where the query's letter appeared, which on a match trial is the target.

\begin{figure}[t]
  \centering
  \includegraphics[width=\textwidth]{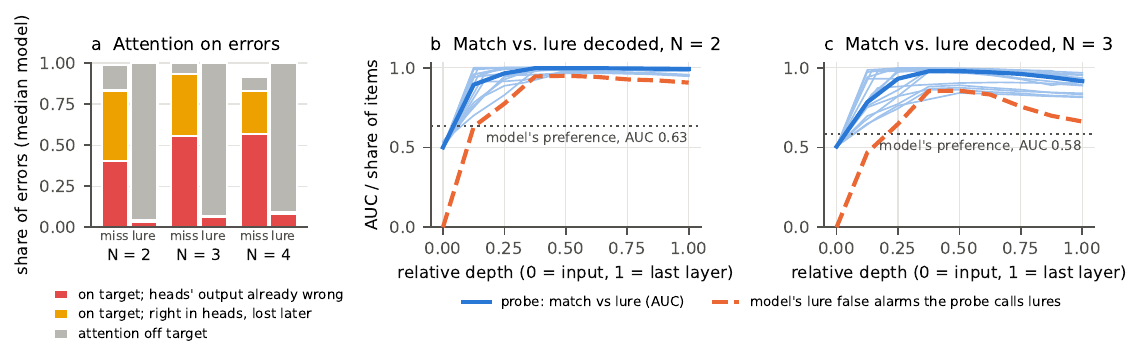}
  \caption{\textbf{LLMs find the target but misread it.} (a) Misses and lure false alarms (FA), split by whether the top heads attended to the target and, if so, whether their output already pointed to the wrong answer (median over models; 14 at N = 2 and 3, 12 at N = 4). Attention is on target for most misses and off target for nearly all lure false alarms. (b, c) A linear probe on the residual stream separates matches from lures (blue; thin lines are individual models, thick line the median) far better than the model's own preference (dotted), and classifies most of the model's own lure false alarms as lures (orange). Depth is scaled from the input (0) to the last layer (1).}
  \label{fig:locate}
\end{figure}

The critical trials are misses: true matches that the model reports as ``no match.'' If misses came from imprecise attention, attention should be off target on them. In fact, the heads attend elsewhere on only 16\%, 7\% and 8\% of misses at N = 2, 3 and 4 (Fig.~\ref{fig:locate}a). To see where the answer goes wrong, we trained a classifier on correct trials to read the answer from the heads' outputs and applied it to the misses. On 41--57\% of misses, the heads' outputs already pointed to the wrong answer although their attention was on the target. On 27--43\%, the outputs pointed to the right answer, which was then lost in later layers. In both groups, attention had found the target.

Lure false alarms show a different pattern (Fig.~\ref{fig:locate}a). On 91--96\% of them a neighboring line, typically the lure's, receives more attention than the target. This is expected of a head that searches by content, because on a lure trial the query's letter is at the lure. The error lies in treating the lure as a match, and avoiding it requires information about whether the matching letter was N$\pm$1 or N items back. We next ask whether the model has this information.

\textbf{The model knows where the match was.} Matches and lures both contain the query's letter; only its position differs. We trained a linear probe, a logistic-regression classifier, to tell matches from lures using the residual stream at the query, separately at nine depths and with training and test sequences kept apart. If such a probe succeeds, the information is present in a form that a single linear unit downstream could read. We score the probe and the model by the area under the ROC curve (AUC); for the model, the score is its preference for ``\texttt{m}'' over ``\texttt{-}''. The AUC is the probability that a random match scores higher than a random lure (0.5 at chance, 1 for perfect separation).

The model's own preference separates matches from lures barely above chance (median AUC 0.63 at N = 2, 0.58 at N = 3; Fig.~\ref{fig:locate}b,c). A probe reading the same model's activations separates them almost perfectly (median 1.00 and 0.98 at the best depth). The probe is at chance at the input (depth 0 in Fig.~\ref{fig:locate}b,c), as it must be, since the query's embedding holds only its own letter; the model's layers compute this information. This information is present even on trials the model gets wrong: the probe classifies a median 95\% (N = 2) and 86\% (N = 3) of the model's lure false alarms as lures. In the largest model, Qwen2.5-14B, the model's own AUC is 0.81 and 0.64, the probe's 1.00 and 0.997, and the probe classifies 99.8\% and 95\% of its lure false alarms as lures. The information peaks in middle layers and fades toward the output, as relative-position information does in \citet{xiong2026superposition}.

The models therefore locate the target and compute where the matching letter was, but their decisions make little use of this information, which places the bottleneck in readout.

\section{Removing interference helps more than sharpening attention}
\label{sec:causal}

\begin{figure}[t]
  \centering
  \includegraphics[width=\textwidth]{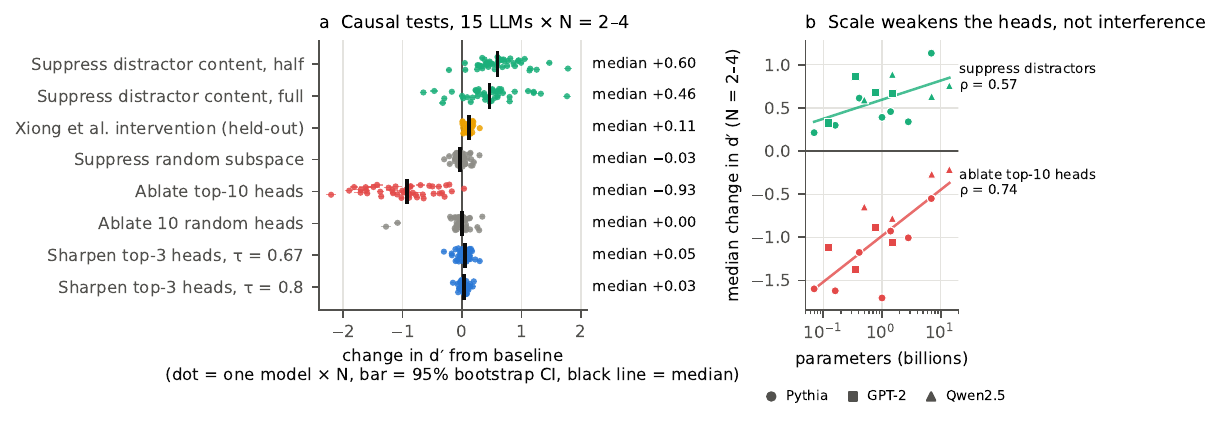}
  \caption{\textbf{Removing interference helps more than sharpening attention.} (a) Change in \dprime{} from the unmodified model under each intervention. Each dot is one model at one N with its 95\% bootstrap interval; black lines are medians. (b) Each model's median change over N = 2--4 against its size, for removing distractor content (half strength) and ablating the top ten heads; lines are least-squares fits against log size.}
  \label{fig:causal}
\end{figure}

The evidence so far is correlational. Heads might attend to the target yet pass on a blurred copy of it that sharper attention would clean up. We therefore intervene on the models. We apply each intervention to all 15 models at N = 2--4 on the same sequences, recalibrating the decision threshold each time (Fig.~\ref{fig:causal}a; Table~\ref{tab:b1}). Each intervention is compared with the unmodified model by a paired bootstrap over sequences (95\% intervals, 2,000 resamples). Two cases whose baseline \dprime{} is below 0.3 are excluded, leaving 43 model-by-N cells.

\textbf{The identified heads are necessary.} Replacing the outputs of the ten most important heads with their average over calibration items, which removes the trial-specific information they carry, lowers \dprime{} in 42 of 43 cells (median $-0.93$; Fig.~\ref{fig:causal}a). Ten random heads from the same layers have no effect (median $+0.00$).

\textbf{Sharper attention barely helps.} Dividing the attention scores of the top three heads by a temperature $\tau<1$ before the softmax concentrates their attention. This changes \dprime{} by a median $+0.05$ at $\tau=0.67$ and $+0.03$ at $\tau=0.8$ (range $-0.30$ to $+0.30$; Fig.~\ref{fig:causal}a). The gain is consistent (Wilcoxon $p=0.04$ and $0.001$) but small. Flattening the same heads ($\tau=2$) costs a median 0.32 (Table~\ref{tab:l5}). When we instead change the temperature of every head at once (N = 2 and 3), \emph{flattening} is best in 19 of 30 cells ($\tau=1.25$ in 16, $\tau=1.5$ in 3) and sharpening in only 2 (Table~\ref{tab:l5}). The heads are therefore already close to their optimal sharpness.

\textbf{Removing distractor content helps substantially.} We first find the directions in the residual stream that encode letter identity: the 19-dimensional subspace spanned by the average activations of the 20 letters (after subtracting their mean), estimated on calibration sequences. Then, for each query, we remove half of each earlier item's component in this subspace, at every layer and over the item's whole line, sparing only the target. This leaves the target and the positions of all items intact but blurs which letter each distractor was. It raises \dprime{} in 42 of 43 cells (median $+0.60$, up to $+1.78$; Fig.~\ref{fig:causal}a), including at N = 4 (median $+0.44$). As the interference account predicts, false alarms fall from about 0.2--0.3 to about 0.1 while hit rates barely change (Table~\ref{tab:l5}). Two controls show that the effect is specific. Removing a random subspace of the same size has no effect (median $-0.03$), and removing the target's letter instead destroys performance (median $-1.54$; Table~\ref{tab:l5}). Complete removal also helps (median $+0.46$) but over-suppresses in Qwen2.5-7B. This intervention requires knowing the target, so it serves only as a causal test. Replicating the single-direction removal of \citet{xiong2026superposition} at answer positions, with its configuration chosen on calibration sequences, gives a median $+0.11$ (28 cells; $+0.12$ with selection on test trials), about a fifth as much. Their gains were largest in the weakest models; ours grow with model size (Fig.~\ref{fig:causal}b). Distractor suppression and head ablation both differ from zero with Wilcoxon $p<10^{-8}$.

\textbf{Why sharpening can do so little.} A simple model of the readout explains this ordering. The quantity the model decides on contains the signal (does the query match the target?) plus the letter content of other items, written in by many heads in many layers. Sharpening one head removes only the distractor content that this head contributes, which is small when the head already attends mostly to the target. Removing distractor content, by contrast, works on all of it, whichever head wrote it.

\begin{proposition}[sharpening versus removing interference]\label{prop:interference}
Let the decision variable be $r=w^\top h$ with $h=\gamma\,y\,e+\sum_{j\in D}B_j g_j+z$, where $y=\pm1$ marks match and non-match, $\gamma e$ is the signal, $g_j\in\mathbb{R}^d$ is the letter content of distractor item $j\in D$, which lies in a $q$-dimensional subspace, $B_j$ are fixed linear maps that write it into $h$, and $z$ is noise independent of the letters. Distractor content may depend on $y$ (lures occur on non-match trials): let $b_y=\mathbb{E}[w^\top\sum_jB_jg_j\mid y]$, $\beta=b_{+}-b_{-}$, and let $\sigma_D^2$ and $\sigma_z^2$ be the within-class variances of the distractor and noise terms of $r$. Then $d'=(2\gamma\,w^\top e+\beta)/\sqrt{\sigma_D^2+\sigma_z^2}$.
\begin{enumerate}\setlength{\itemsep}{1pt}\setlength{\topsep}{2pt}
  \item[(a)] Removing a fraction $\alpha$ of every distractor's letter content ($g_j\to(1-\alpha)g_j$) gives $d'(\alpha)=(2\gamma\,w^\top e+(1-\alpha)\beta)/\sqrt{(1-\alpha)^2\sigma_D^2+\sigma_z^2}$, which increases with $\alpha$ when $\beta\le0$, that is, when distractors push non-match trials toward ``match'', as lures do.
  \item[(b)] Removing a uniformly random $q$-dimensional subspace instead removes, in expectation, a fraction $q/d$ of each distractor's content (in squared norm).
  \item[(c)] If a head with value vectors $v_j$ and output $W_O\sum_j a_j v_j$ reaches $h$ directly and puts attention weight at least $1-\epsilon$ on the target line, it contributes at most $\epsilon\max_j|w^\top W_O v_j|$ of distractor content to $r$; this bounds what re-weighting its attention can remove.
\end{enumerate}
\end{proposition}

The proposition's predictions match the data: suppression lowers false alarms while sparing hits, by removing both the lure bias $\beta$ and the variance $\sigma_D^2$; a random subspace removes almost nothing ($q/d\le19/512$); and sharpening is bounded by the small off-target share of a few heads. The model does not explain why half-strength removal beats full removal. Proofs are in Appendix~\ref{app:theory}.

\textbf{Effects of scale.} The effects of the two interventions diverge as models grow (Fig.~\ref{fig:causal}b). Ablating the identified heads costs less in larger models (Spearman $\rho=0.74$ between size and each model's median change, $p=0.002$, 15 models), from about $-1.6$ in the smallest Pythia models to $-0.21$ in Qwen2.5-14B, indicating that the solution becomes more distributed across heads. In contrast, removing distractor content helps more in larger models ($\rho=0.57$, $p=0.03$). Qwen2.5-14B shows this most clearly. Half-strength suppression raises its \dprime{} from 2.42 to 2.88, from 1.60 to 3.38 and from 1.29 to 2.05 at N = 2, 3 and 4, with every bootstrap interval at least 0.40 above zero, while a random subspace changes it by $-0.10$, $-0.05$ and $+0.18$. With distractor content removed, the model does better at 3-back than it does unaided at 2-back. Over a 200-fold range of size, then, the heads that implement the task change, while the interference bottleneck remains.

\section{Attention precision in a minimal transformer}
\label{sec:theory}

\begin{figure}[t]
  \centering
  \includegraphics[width=\textwidth]{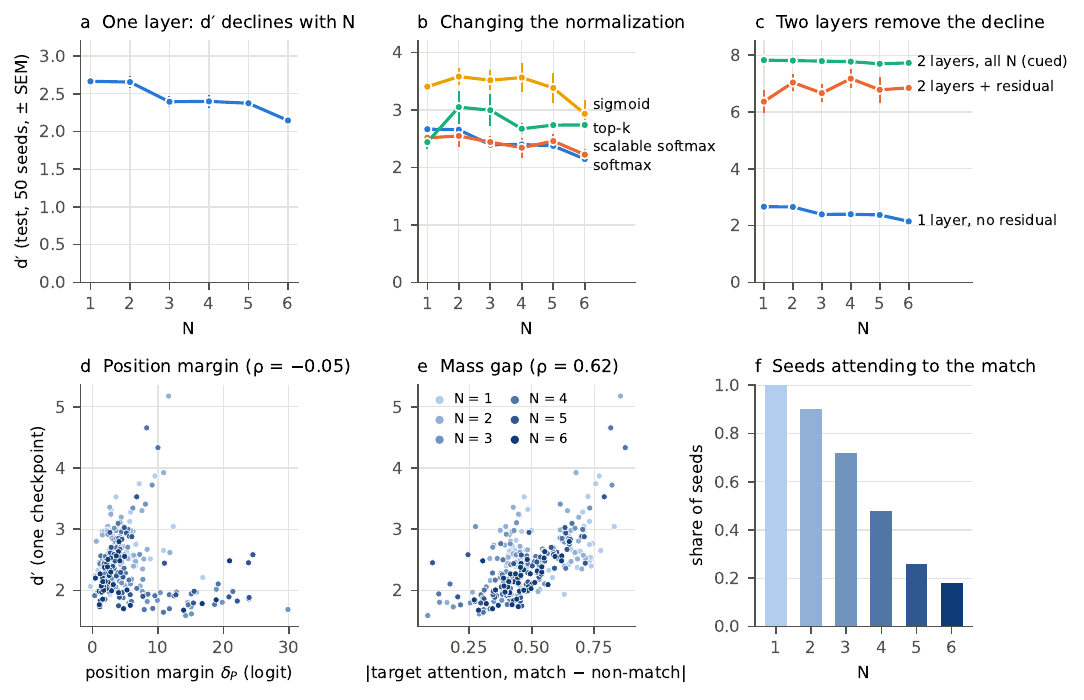}
  \caption{\textbf{Attention precision in minimal transformers trained from scratch.} (a) A one-layer model without residual stream declines with N (50 seeds, $\pm$SEM). (b) Sigmoid and top-$k$ attention flatten the decline; scalable softmax does not. (c) Two layers with a residual stream remove it, both for one model per N and for one cued model trained on all N. (d, e) Across 300 one-layer models (50 seeds $\times$ N = 1--6; darker = larger N), \dprime{} is unrelated to the position margin $\delta_P$ but tracks how much attention to the target differs between match and non-match trials. (f) Share of models that attend more to the target on match trials.}
  \label{fig:minimal}
\end{figure}

The precision account was originally proposed for a minimal transformer: a single attention layer with no residual stream and no further processing, trained on N-back \citep{gong2024selfattention}. Such a model does decline with N (\dprime{} from 2.67 to 2.15 over N = 1--6; Fig.~\ref{fig:minimal}a). We trained this model at each N with 50 random seeds, 300 models in all. We analyze this model exactly, identify what limits it, and test whether the limit persists with a second layer.

\textbf{The minimal model.} For a query at position $i$, the model attends with weights $a_{ij}$ to every position $j$ and outputs $o_i=\sum_j a_{ij}v_j$, where $v_j$ is the value vector of item $j$; the decision is a linear function of $o_i$. Attention scores have a content term and a position term, $s_{ij}=c(x_i,x_j)+p(i-j)$, where $x_j$ is the letter at $j$. Let $t=i-N$ be the target, $u_j=w^\top v_j$ the amount item $j$ adds to the decision, and $r_i=\sum_j a_{ij}u_j$ the decision variable. Without a residual stream, $u_j=u(x_j,j)$ depends only on the letter and position of item $j$. Our sequences treat all letters alike, so every single position has the same letter distribution on match and non-match trials; only the joint distribution of letters differs.

\textbf{A one-layer model can detect a match only through where it attends.} The output is a weighted average of letter vectors, and a weighted average with fixed weights cannot tell whether two letters are equal. The comparison must therefore happen in the attention weights themselves.

\begin{proposition}[the readout is a test of attention mass]\label{prop:mass}
In this model the decision is linear in $\sum_j a_{ij} v_j$, and for fixed weights $a_{ij}$ no linear function of a convex combination of letter embeddings can test whether two letters are equal. The difference in mean readout between match and non-match trials must therefore be carried by how the attention distribution differs between them.
\end{proposition}

\begin{corollary}[precise location without content carries no signal]\label{cor:fixed}
For any fixed weight vector $\bar a$ and $u_{\max}=\max_{x,j}|u(x,j)|$,
\[
\bigl|\mathbb{E}[r_i\mid\text{match}]-\mathbb{E}[r_i\mid\text{non-match}]\bigr|\;\le\;u_{\max}\bigl(\mathbb{E}_{\text{match}}\|a_i-\bar a\|_1+\mathbb{E}_{\text{non-match}}\|a_i-\bar a\|_1\bigr).
\]
In particular, any attention pattern that does not depend on content, including one that splits all weight between $i$ and $i-N$, gives the same mean readout on match and non-match trials, however precisely it locates the target.
\end{corollary}

Both statements hold in trained models. We took these 300 models, together with 420 variants that differ in width, amount of training data or weight decay (720 one-layer models in all), and replaced their attention by a fixed mix of positions $i$ and $i-N$, so that both letters reach the output: \dprime{} stays within $\pm0.07$ of zero at every N. Removing the content term from the attention scores gives the same result ($|d'|\le0.04$). Precise location is therefore uninformative unless attention also responds to content. Performance is instead predicted by how much attention to the target \emph{changes} between match and non-match trials: across the original 300 models, the size of this change correlates with \dprime{} at Spearman $\rho=0.62$ (Fig.~\ref{fig:minimal}e). Its sign changes with N. At N = 1 every model attends more to the target when it matches; at N = 4 only 48\% do and at N = 6 only 18\%, while the rest attend \emph{away} from a matching letter (Fig.~\ref{fig:minimal}f). Both strategies test where the attention mass goes, and Corollary~\ref{cor:fixed} constrains them alike.

\textbf{The role of softmax normalization.} The precision account rests on a property of softmax: competing positions dilute the target's weight.

\begin{proposition}[softmax dispersion]\label{prop:dispersion}
If each of the $K$ other positions has a score at least $\Delta$ below the target's, the target receives weight at least $1/(1+Ke^{-\Delta})$, and this guarantees weight $1-\epsilon$ for every such configuration if and only if $\Delta\ge\log(K(1-\epsilon)/\epsilon)$. Two positions with the same letter are separated only by the position term: an item at distance $N\pm1$ receives $e^{-\delta_P}$ times the weight of an item at distance $N$, where $\delta_P=p(N)-p(N\pm1)$ is the \emph{position margin}.
\end{proposition}
If dispersion caused the decline, lure false alarms should fall as $\delta_P$ grows and $\delta_P$ should shrink with N (Corollary~\ref{cor:lure}, Appendix~\ref{app:theory}). Changing the normalization does change the decline. Replacing softmax by unnormalized sigmoid attention, or by attention restricted to the top $k$ scores, raises and flattens the curve, whereas scalable softmax \citep{nakanishi2025scalable}, designed to counter dispersion over long contexts, does not (Fig.~\ref{fig:minimal}b). The specific predictions of Corollary~\ref{cor:lure}, however, both fail. When we decompose the attention scores of the original 300 models, the position margin $\delta_P$ does not fall steadily with N (4.5 at N = 1, rising to 6.7 at N = 5, 3.6 at N = 6), and across these models it predicts neither lure false alarms ($\rho=-0.03$) nor \dprime{} ($\rho=-0.05$; Fig.~\ref{fig:minimal}d). Sharpening the trained attention also lowers \dprime{} at N = 1--5 (from 2.70 to 2.20 at N = 1 with $\tau=0.67$), because the weight on other positions is part of how these models compute the answer. Precision thus bounds the minimal model, but its decline with N reflects how attention responds to content rather than how sharply it codes position.

\textbf{A second layer removes the limit.} With two layers, a first layer can copy the letter from N back into the query's position; a second layer, or the feed-forward block (MLP) that follows attention, can then compare the two letters directly, without competing against all other items.

\begin{proposition}[depth removes the bound]\label{prop:depth}
With two layers, a residual stream and an absolute or rotary position code, a first-layer head can copy $x_{i-N}$ into position $i$ with a margin that grows with the weight norm. Position $i$ then carries both letters, and a second layer (through its query--key product) or an MLP can test their equality without competing against the other, N-dependent positions. A position bias that decreases monotonically with distance (ALiBi, or no position code) cannot single out distance $N>1$ in one head, so the copy step fails.
\end{proposition}

Trained models match these predictions (Fig.~\ref{fig:minimal}c). As Proposition~\ref{prop:depth} predicts, and as \citet{xiong2026superposition} found with rotary codes, the decline disappears with two layers and a residual stream: \dprime{} is 6.4--7.2 at every N with learned position codes, 5.4--5.8 with sinusoidal and 4.5--5.9 with rotary codes \citep{su2021roformer}, and 7.7--7.8 for a single model trained on all N with a cue. As predicted, models with ALiBi \citep{press2022alibi} or no position code fail at every N (\dprime{} 1.3--1.8); their failure at N = 1 goes beyond the prediction. Table~\ref{tab:predictions} lists all predictions and tests.

Precision therefore limits only models too shallow to compare two letters directly. The pretrained LLMs we tested have 6 to 48 layers, and Sections~\ref{sec:locate} and~\ref{sec:causal} show that their limit lies in readout.

\section{Related work}
\label{sec:related}

\textbf{Working memory in LLMs.} \citet{gong2024chatgpt} found human-like N-back declines in LLMs, and \citet{gong2024selfattention} related the decline in small trained transformers to the spread of attention. \citet{hu2025nback} showed that weaker models drift to an easier task, which we control for; \citet{zhang2024wmreasoning} relate working-memory tests to reasoning. Closest to us, \citet{xiong2026superposition} show that LLMs fail an N-back task that a two-layer transformer solves, find behavioral and representational signatures of interference, and improve performance by suppressing letter-identity directions. We add a direct test of the two accounts on the trials the models get wrong, and the conditions under which precision does bind.

\textbf{Convergent representations and mechanisms.} Representations of different networks, and of networks and brains, are often compared with similarity measures such as representational similarity analysis and CKA \citep{kriegeskorte2008rsa,kornblith2019similarity,sucholutsky2023alignment}, and larger models have been argued to converge on shared representations \citep{huh2024platonic}. Mechanistic studies ask the same question of circuits, finding features and algorithms that recur across models \citep{olah2020zoom,chughtai2023toy,gurnee2024universal}. It is also well established that decodable information need not be used \citep{dewit2016neuroimaging,hewitt2019control,ravichander2021probing,elazar2021amnesic}. Our study compares a mechanism across 15 models and with accounts of human working memory.

\textbf{Attention limits and mechanisms.} Softmax attention disperses as the number of positions grows \citep{velickovic2025softmax,hayase2026critical}, which scalable softmax \citep{nakanishi2025scalable} and sigmoid attention \citep{ramapuram2024sigmoid} change, and expressivity results bound what one attention layer can compute \citep{hahn2020limitations,sanford2023representational}. Our analyses build on circuit studies \citep{elhage2021framework,wang2023ioi}, induction and retrieval heads \citep{olsson2022induction,wu2024retrieval}, and activation and attribution patching \citep{zhang2024patching,kramar2024atp}.

\section{Discussion}
\label{sec:discussion}

We asked why LLMs, which keep every item in context, show a working-memory limit similar to that of humans. The limit does not come from a failure to find the item. The heads that carry the task land on the target even when the model errs, and the model's activations record where a matching letter appeared even when it falls for a lure. The limit arises in readout, where the content of other items interferes with the decision. Removing that content restores much of the lost performance, whereas sharpening attention has little effect. The precision account does describe a real limit, but only for transformers too shallow to compare two items directly.

These results bear on when different networks converge. Fifteen models from three families, trained on different data at sizes spanning a 200-fold range, show the same dissociation between locating the target and reading it out. Only the implementation varies: larger models spread the task over more heads and gain more from removing interference. Like \citet{xiong2026superposition}, we find that this shared mechanism resembles interference accounts of human working memory \citep{oberauer2012interference,szmalec2011interference,engle2002executive}. On these accounts, items are available but compete at the moment of use, and lure errors arise from that competition rather than from lost information. Our results add a distinction that is hard to test in people. A wrong answer could come from retrieving the wrong item or from misusing the right one, and behavior alone cannot tell these apart. In the models we can observe both steps on every trial, and find that using an item can fail even when finding it succeeds.

The gap between what a probe reads from the models' activations and what their decisions use also cautions against comparing systems by decodable content alone. Decodable information need not drive behavior \citep{dewit2016neuroimaging,ravichander2021probing}, so two systems can appear similar or dissimilar to a decoder for reasons unrelated to their behavior. Interventions on the candidate content, as in Section~\ref{sec:causal}, can tell these cases apart. The decline is not specific to softmax attention: a selective state-space model trained from scratch also declines with N (Appendix~\ref{app:arch}). For improving working memory in models, these results suggest that separating item content from the decision is more promising than sharpening attention.

\textbf{Limitations.} We study one task family, letter N-back with synthetic stimuli, and the mechanistic analyses use the line-by-line format with correct earlier answers shown. The head-level account may not transfer beyond 14B, since the identified heads matter less in larger models, although the benefit of removing interference grows over our range. The probes show only that the information is linearly decodable, and in the GPT-2 family at N = 3 part of it is lost by the last layer (probe AUC 0.84--0.89). Propositions~\ref{prop:mass}--\ref{prop:depth} concern the minimal model; Proposition~\ref{prop:interference} is a linear model of the readout, not derived from trained weights.

\newpage
\bibliographystyle{plainnat}
\bibliography{refs}

\clearpage
\appendix

\section{Proofs}
\label{app:theory}

Table~\ref{tab:predictions} lists each theoretical prediction and the experiment that tests it.

\begin{table}[h]
  \centering\small
  \caption{Predictions of the propositions and their tests. \checkmark: confirmed; $\times$: not supported.}
  \label{tab:predictions}
  \begin{tabular}{@{}>{\raggedright\arraybackslash}p{0.095\textwidth}>{\raggedright\arraybackslash}p{0.34\textwidth}>{\raggedright\arraybackslash}p{0.25\textwidth}>{\raggedright\arraybackslash}p{0.215\textwidth}@{}}
    \toprule
    Statement & Prediction & Test & Result \\
    \midrule
    Prop.~\ref{prop:interference} & Removing distractor content beats sharpening, cuts false alarms, keeps hits; random subspace $\approx0$ & 15 LLMs, N = 2--4 & $+0.60$ vs.\ $+0.05$; random $-0.03$ \checkmark \\
    Cor.~\ref{cor:fixed} & Content-independent attention gives no signal, however precise & Fixed attention on $i$, $i-N$; content term removed & $|d'|\le0.07$ \checkmark \\
    Prop.~\ref{prop:mass} & \dprime{} tracks the match/non-match attention gap & 300 one-layer models & $\rho=0.62$ \checkmark \\
    Prop.~\ref{prop:dispersion} & Changing the normalization changes the decline & Sigmoid, top-$k$, scalable softmax & sigmoid, top-$k$ flatten it \checkmark \\
    Cor.~\ref{cor:lure} & Lure false alarms fall with $\delta_P$; $\delta_P$ shrinks with N & Attention-score decomposition & $\rho=-0.03$; $\delta_P$ not monotone in N $\times$ \\
    Prop.~\ref{prop:depth} & Two layers remove the decline with absolute or rotary codes & Depth and position-code variants & flat, \dprime{} 4.5--7.8 \checkmark \\
    \bottomrule
  \end{tabular}
\end{table}

\textbf{Proposition~\ref{prop:interference}.} The class means of $r$ are $\pm\gamma\,w^\top e+b_{\pm}$ (plus a common constant), so their difference is $2\gamma\,w^\top e+\beta$, and the within-class variance is $\sigma_D^2+\sigma_z^2$. (a) The content enters $h$ linearly, so scaling every $g_j$ by $1-\alpha$ scales $\beta$ by $1-\alpha$ and $\sigma_D^2$ by $(1-\alpha)^2$; $\gamma$, $e$ and $z$ do not involve distractor letters. With $\beta\le0$ and a positive numerator, the numerator rises and the denominator falls as $\alpha$ grows. (b) For a uniformly random $q$-dimensional subspace with projector $P_R$, $\mathbb{E}[P_R]=(q/d)I$ by rotational symmetry, so $\mathbb{E}\|P_Rg\|^2=g^\top\mathbb{E}[P_R]g=(q/d)\|g\|^2$. (c) The head's distractor contribution is $\sum_{j\in D}a_jw^\top W_Ov_j$, and $\sum_{j\in D}a_j\le\epsilon$ because the distractors lie off the target line. \qed

\textbf{Proposition~\ref{prop:mass}.} Let $e_x$ be the value vector of letter $x$ (no residual stream, no MLP, so the output is $o_i=\sum_j a_{ij}e_{x_j}$ up to a fixed linear map, plus positional components that do not depend on the letters). A linear readout computes $w^\top o_i=\sum_j a_{ij}w^\top e_{x_j}$. For fixed attention weights this is additive in the letters, so it cannot represent the indicator $[x_i=x_{i-N}]$, which is not additive (it is an XOR-like function of the two letters). Any dependence of the decision on equality must enter through $a_{ij}$, which depends on $x_i$ and $x_j$ jointly through the content term $c(x_i,x_j)$. \qed

\textbf{Corollary~\ref{cor:fixed}.} Write $r_i=\sum_j\bar a_ju_j+\sum_j(a_{ij}-\bar a_j)u_j$. The first term is a sum of functions of single letters, $\bar a_j u(x_j,j)$. Because the generator is invariant under relabeling the alphabet, each $x_j$ is uniform on both match and non-match trials, so the first term has the same expectation on both. The second term is at most $u_{\max}\|a_i-\bar a\|_1$ in absolute value. Taking expectations on both trial types gives the bound. If attention does not depend on content, $a_i$ is fixed for each $i$; with $\bar a=a_i$ the bound is zero, so the means are equal. The distributions need not be equal (joint letter statistics differ between trial types), so a threshold can retain a small sensitivity. \qed

\textbf{Proposition~\ref{prop:dispersion}.} With target score $s_t$ and $K$ competitors $s_k\le s_t-\Delta$, $a_t=e^{s_t}/(e^{s_t}+\sum_k e^{s_k})\ge 1/(1+Ke^{-\Delta})$. Requiring $a_t\ge1-\epsilon$ gives $Ke^{-\Delta}\le\epsilon/(1-\epsilon)$, so $\Delta\ge\log(K(1-\epsilon)/\epsilon)$. For a lure $\ell$ with $x_\ell=x_t$, $c(x_i,x_\ell)=c(x_i,x_t)$ and $a_\ell/a_t=\exp(p(i-\ell)-p(i-t))=e^{-\delta_P}$. \qed

\begin{corollary}[what the precision account predicts]\label{cor:lure}
Suppose the decision is driven by the weight on an item that carries the query's letter. Compare a match trial with a lure trial that is identical except that the query's letter sits at distance $N\pm1$ instead of $N$. The lure then receives $e^{-\delta_P}Z^{\text{m}}/Z^{\text{l}}$ times the weight the target receives on the match trial, where $Z^{\text{m}},Z^{\text{l}}$ are the two softmax denominators. Unless the denominators compensate, lure false alarms should fall as $\delta_P$ grows, and if dispersion caused the decline with N, $\delta_P$ should shrink as N grows.
\end{corollary}

\textbf{Corollary~\ref{cor:lure}.} On the match trial the target carries the query letter and receives $a_t=e^{c^+ +p(N)}/Z^{\text{m}}$, where $c^+=c(x,x)$. On the paired lure trial the lure carries it and receives $a_\ell=e^{c^+ +p(N\pm1)}/Z^{\text{l}}$. Hence $a_\ell/a_t=e^{-\delta_P}\,Z^{\text{m}}/Z^{\text{l}}$. The two denominators differ only in the terms of these two positions, so the ratio is near 1 when those terms are a small share of either; when the matching item dominates $Z^{\text{m}}$ the ratio exceeds 1 and weakens the prediction. \qed

\textbf{Proposition~\ref{prop:depth}} (sketch). With absolute positions $P_j$ and a first-layer head whose query--key product is $\beta\,P_i^\top R P_j$ with $R$ chosen so that $P_i^\top RP_j$ is maximized uniquely at $j=i-N$ with gap $g>0$, the head places weight at least $1/(1+Le^{-\beta g})$ on $i-N$, where $L\le i$ is the number of positions; its output, added to the residual stream, makes $e_{x_{i-N}}$ available at position $i$ alongside $e_{x_i}$, and an MLP or a second attention layer can compute equality of the two letters. Rotary codes provide the same relative-offset selectivity. Under ALiBi, the bias $-\lambda(i-j)$ is monotone in distance, so one head's maximum over positions with equal content is always the most recent, and distance $N>1$ cannot be singled out. \qed

\section{Methods}
\label{app:methods}

\textbf{Minimal models.} Attention-only transformers ($d_{\text{model}}=512$ for the main one-layer configuration, 128 for variants), Adam with learning rate $10^{-4}$, batch 32, fresh sequences each step, up to 30,000 steps with early stopping on validation loss (patience 2,000); 50 seeds for the base configuration and 10 for each variant. Variants: depth and residual stream; learned, sinusoidal, rotary, ALiBi and no position code; softmax, scalable softmax, sigmoid and top-$k$ attention. The cued multi-N model is a two-layer residual model with a cue token giving N. Recurrent baselines (LSTM, LRU, linear attention, selective SSM) use state sizes 32, 128 and 512. The 720 one-layer checkpoints used for post-hoc attention edits are the base configuration (50 seeds) and seven variants in width, training-set size and weight decay (10 seeds each), at N = 1--6; the attention-score decomposition and the correlations use the base configuration alone (300 checkpoints).

\textbf{LLM evaluation.} Prompts start with a one-sentence instruction and ``N = $n$'', then three demonstration sequences with correct answers and the test sequence. Letters, ``\texttt{ m}'' and ``\texttt{ -}'' are single tokens in every tokenizer (checked per model). Models up to 1B run in float32 and larger models in bfloat16. The chat format, used for Qwen2.5-Instruct models, sends one user turn per letter and one assistant turn per answer. The best-fitting $k$ is the $k$-back rule whose labels best match the model's answers.

\textbf{Head ranking.} Clean and corrupted prompts differ only in the letter at $i-N$, flipping the answer. For each head we patch its output from the clean run into the corrupted run at all positions and report the recovered fraction of the mean logit-difference gap (set to missing when the gap is below 0.1).

\textbf{Locating analysis.} Attention from the query letter token to all tokens of a line is summed per head and averaged over the top five heads. The answer classifier is a logistic regression on the top five heads' outputs at the query, trained on up to 4,000 correct items and applied to errors.

\textbf{Probe.} Items with a same-letter item at $i-N$ (matches) or at a lure distance (lures); logistic regression with $L_2$ penalty on standardized features, five folds split by sequence.

\textbf{Interventions.} Sharpening multiplies a head's attention scores by $1/\tau$; ablation replaces a head's output by its mean over calibration items; random heads are drawn from the same layers. Suppression replaces $h$ by $h-\alpha\,((h-\mu)U)U^\top$ at every residual site (the input to the first layer and every layer's output) over the tokens of each non-target line, where $U$ spans the 19 centerd class means of the 20 letters at test-item letter tokens. The random control uses a random orthonormal $U$ of the same rank.

\textbf{Compute.} Minimal models mostly on a CPU cluster (about 1,060 core-hours), with some sweeps on single H100 GPUs; LLM studies on single L40S, H100 or B200 GPUs per model and N. The whole project, including preliminary and failed runs, used about \gpuhours{} GPU-hours.

\textbf{Broader impact.} This is basic research on model mechanisms with public models and synthetic data. Better diagnosis of working-memory failures may help evaluation; the interventions are analysis tools with no deployment use.

\textbf{Licenses.} Pythia (Apache-2.0), GPT-2 (MIT), Qwen2.5 (Apache-2.0; Qwen license for some sizes).

\section{Other architectures and development}
\label{app:arch}

\begin{figure}[h]
  \centering
  \includegraphics[width=\textwidth]{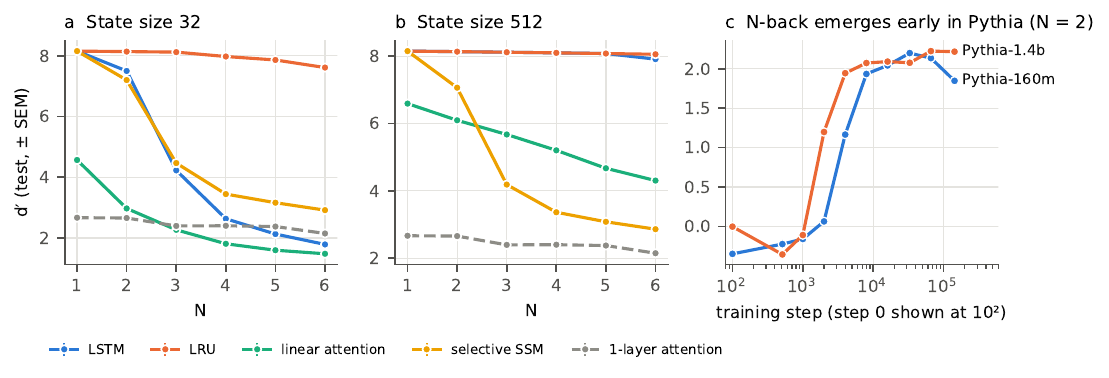}
  \caption{\textbf{Other architectures and development.} (a, b) Recurrent and linear-attention baselines at state sizes 32 and 512, with the one-layer attention model (dashed) for reference; LSTM-512 overlaps LRU-512. (c) N-back \dprime{} over Pythia training (N = 2).}
  \label{fig:arch}
\end{figure}

Recurrent baselines trained on the same stimuli separate two kinds of memory (Fig.~\ref{fig:arch}a,b). LSTM \citep{hochreiter1997lstm} and LRU \citep{orvieto2023lru} networks with a 512-dimensional state solve every N (\dprime{} $\ge7.9$); with a 32-dimensional state the LSTM declines (8.2 to 1.8) while the LRU barely does. Linear attention \citep{katharopoulos2020linear} declines at both sizes, and a selective state-space model \citep{gu2023mamba}, whose gating depends on the input as attention does, declines from 8.2 to 2.9 even with a 512-dimensional state. The decline is therefore not specific to softmax attention and also appears in architectures that retrieve by content-dependent weighting \citep{arora2023zoology,jelassi2024repeat}. In Pythia, the top N-back heads and the behavior appear early in training: \dprime{} at N = 2 rises from chance to near its final value between steps 1,000 and 8,000 (Fig.~\ref{fig:arch}c).

\section{Additional results}
\label{app:results}

\begin{table}[h]
  \centering\small
  \caption{\dprime{} by model and N (line-by-line format, correct earlier answers shown), and false-alarm rates on N$\pm$1 lures versus other non-matches at N = 2.}
  \label{tab:l2}
  \begin{tabular}{lrrrrrrr}
\toprule
Model & $N{=}1$ & 2 & 3 & 4 & 5 & 6 & lure FA, $N{=}2$ \\
\midrule
Pythia-70m & 1.25 & 1.52 & 1.40 & 0.03 & 0.45 & -0.06 & 0.83 / 0.12 \\
Pythia-160m & 2.10 & 1.85 & 1.92 & 0.96 & 0.79 & 0.10 & 0.79 / 0.02 \\
Pythia-410m & 2.37 & 1.76 & 1.25 & 0.53 & 0.56 & 0.44 & 0.76 / 0.04 \\
Pythia-1b & 2.57 & 2.06 & 1.46 & 0.64 & 0.73 & 0.54 & 0.83 / 0.04 \\
Pythia-1.4b & 3.03 & 2.21 & 1.36 & 0.99 & 0.74 & 0.80 & 0.73 / 0.02 \\
Pythia-2.8b & 2.71 & 2.10 & 1.42 & 0.94 & 0.87 & 0.63 & 0.82 / 0.03 \\
Pythia-6.9b & 3.19 & 1.93 & 1.26 & 1.07 & 1.08 & 1.00 & 0.64 / 0.03 \\
GPT-2 & 2.15 & 2.08 & 1.28 & 0.87 & 0.73 & -0.02 & 0.96 / 0.08 \\
GPT-2 medium & 2.27 & 1.92 & 1.35 & 0.95 & 0.89 & 0.50 & 0.92 / 0.06 \\
GPT-2 large & 2.36 & 2.24 & 1.43 & 0.91 & 0.67 & 0.21 & 0.95 / 0.04 \\
GPT-2 xl & 2.34 & 1.48 & 0.93 & -0.01 & 0.16 & 0.15 & 0.70 / 0.06 \\
Qwen2.5-0.5B & 2.58 & 1.64 & 1.04 & 0.64 & 0.52 & 0.52 & 0.81 / 0.06 \\
Qwen2.5-1.5B & 2.86 & 2.18 & 1.75 & 1.18 & 1.26 & 0.76 & 0.65 / 0.01 \\
Qwen2.5-7B & 3.17 & 1.53 & 1.44 & 0.88 & 0.61 & 0.79 & 0.64 / 0.06 \\
Qwen2.5-14B & 4.28 & 2.42 & 1.60 & 1.29 & 1.41 & 1.15 & 0.63 / 0.01 \\
\bottomrule
\end{tabular}

\end{table}

\begin{table}[h]
  \centering\small
  \caption{Change in \dprime{} under each intervention; bold entries have 95\% bootstrap intervals excluding zero. $^\dagger$ degenerate baseline (excluded from summaries).}
  \label{tab:b1}
  \begin{tabular}{lrrrrrrr}
\toprule
Model & $N$ & base $d'$ & sharpen & ablate top & ablate rand. & suppress & supp. rand. \\
\midrule
Pythia-70m & 2 & 1.52 & \textbf{-0.11} & \textbf{-1.71} & \textbf{-1.08} & \textbf{+0.38} & \textbf{-0.30} \\
 & 3 & 1.40 & \textbf{+0.22} & \textbf{-1.49} & \textbf{-1.28} & \textbf{+0.05} & \textbf{-0.21} \\
 & 4 & 0.03$^\dagger$ & \textbf{+0.97} & \textbf{-0.17} & +0.04 & \textbf{+0.82} & \textbf{+0.31} \\
\addlinespace[2pt]
Pythia-160m & 2 & 1.85 & \textbf{+0.14} & \textbf{-1.62} & +0.01 & \textbf{+1.25} & \textbf{+0.29} \\
 & 3 & 1.92 & \textbf{-0.17} & \textbf{-1.90} & \textbf{-0.13} & \textbf{+0.30} & +0.03 \\
 & 4 & 0.96 & \textbf{+0.30} & \textbf{-1.09} & \textbf{-0.20} & \textbf{-0.22} & +0.01 \\
\addlinespace[2pt]
Pythia-410m & 2 & 1.76 & \textbf{+0.17} & \textbf{-1.36} & \textbf{-0.03} & \textbf{+0.61} & -0.01 \\
 & 3 & 1.25 & \textbf{-0.08} & \textbf{-1.18} & \textbf{+0.13} & \textbf{+0.60} & \textbf{-0.04} \\
 & 4 & 0.53 & \textbf{+0.13} & \textbf{-0.39} & \textbf{+0.08} & \textbf{+0.70} & \textbf{+0.05} \\
\addlinespace[2pt]
Pythia-1b & 2 & 2.06 & \textbf{-0.04} & \textbf{-1.81} & \textbf{+0.13} & \textbf{+0.59} & \textbf{-0.05} \\
 & 3 & 1.46 & \textbf{+0.11} & \textbf{-1.70} & \textbf{+0.34} & \textbf{+0.26} & \textbf{-0.12} \\
 & 4 & 0.64 & \textbf{+0.09} & \textbf{-0.83} & \textbf{+0.29} & \textbf{+0.39} & \textbf{-0.09} \\
\addlinespace[2pt]
Pythia-1.4b & 2 & 2.21 & \textbf{+0.08} & \textbf{-1.61} & \textbf{-0.06} & \textbf{+0.95} & \textbf{-0.05} \\
 & 3 & 1.36 & \textbf{+0.05} & \textbf{-0.93} & \textbf{+0.05} & \textbf{+0.46} & \textbf{+0.05} \\
 & 4 & 0.99 & \textbf{+0.05} & \textbf{-0.66} & +0.02 & \textbf{+0.44} & \textbf{-0.07} \\
\addlinespace[2pt]
Pythia-2.8b & 2 & 2.10 & \textbf{-0.05} & \textbf{-1.59} & \textbf{-0.12} & \textbf{+0.25} & \textbf{-0.03} \\
 & 3 & 1.42 & \textbf{+0.12} & \textbf{-1.01} & \textbf{-0.03} & \textbf{+0.34} & -0.01 \\
 & 4 & 0.94 & \textbf{+0.12} & \textbf{-0.70} & \textbf{+0.02} & \textbf{+0.41} & \textbf{+0.05} \\
\addlinespace[2pt]
Pythia-6.9b & 2 & 1.93 & \textbf{+0.03} & \textbf{-1.02} & \textbf{-0.08} & \textbf{+0.94} & \textbf{-0.08} \\
 & 3 & 1.26 & \textbf{+0.10} & \textbf{-0.34} & \textbf{+0.26} & \textbf{+1.47} & \textbf{+0.13} \\
 & 4 & 1.07 & \textbf{+0.07} & \textbf{-0.55} & +0.03 & \textbf{+1.13} & \textbf{+0.09} \\
\addlinespace[2pt]
GPT-2 & 2 & 2.08 & \textbf{-0.30} & \textbf{-2.20} & \textbf{-0.09} & \textbf{+0.05} & \textbf{-0.22} \\
 & 3 & 1.28 & -0.00 & \textbf{-1.12} & \textbf{-0.09} & \textbf{+0.71} & \textbf{-0.07} \\
 & 4 & 0.87 & \textbf{-0.02} & \textbf{-0.86} & \textbf{+0.07} & \textbf{+0.32} & \textbf{-0.12} \\
\addlinespace[2pt]
GPT-2 medium & 2 & 1.92 & +0.02 & \textbf{-1.38} & \textbf{+0.14} & \textbf{+1.05} & +0.00 \\
 & 3 & 1.35 & \textbf{-0.06} & \textbf{-1.43} & \textbf{-0.10} & \textbf{+0.86} & \textbf{-0.08} \\
 & 4 & 0.95 & \textbf{-0.07} & \textbf{-0.98} & \textbf{-0.18} & \textbf{+0.34} & \textbf{-0.12} \\
\addlinespace[2pt]
GPT-2 large & 2 & 2.24 & \textbf{-0.17} & \textbf{-1.16} & \textbf{-0.09} & \textbf{+0.68} & \textbf{+0.06} \\
 & 3 & 1.43 & \textbf{+0.15} & \textbf{-0.89} & \textbf{+0.02} & \textbf{+0.77} & \textbf{-0.03} \\
 & 4 & 0.91 & \textbf{+0.06} & \textbf{-0.50} & \textbf{-0.05} & \textbf{+0.55} & \textbf{-0.03} \\
\addlinespace[2pt]
GPT-2 xl & 2 & 1.46 & \textbf{+0.12} & \textbf{-1.24} & \textbf{-0.06} & \textbf{+0.52} & \textbf{+0.15} \\
 & 3 & 0.94 & \textbf{+0.12} & \textbf{-0.88} & +0.01 & \textbf{+0.81} & \textbf{-0.05} \\
 & 4 & -0.02$^\dagger$ & \textbf{+0.55} & \textbf{-0.16} & \textbf{+0.53} & \textbf{+0.91} & \textbf{+0.48} \\
\addlinespace[2pt]
Qwen2.5-0.5B & 2 & 1.64 & \textbf{-0.03} & \textbf{-0.87} & \textbf{+0.06} & \textbf{+0.49} & \textbf{+0.15} \\
 & 3 & 1.04 & \textbf{+0.05} & \textbf{-0.65} & \textbf{-0.11} & \textbf{+0.67} & \textbf{-0.07} \\
 & 4 & 0.64 & -0.01 & \textbf{-0.38} & +0.02 & \textbf{+0.60} & \textbf{-0.09} \\
\addlinespace[2pt]
Qwen2.5-1.5B & 2 & 2.18 & \textbf{+0.10} & \textbf{-0.76} & \textbf{+0.06} & \textbf{+0.89} & \textbf{-0.11} \\
 & 3 & 1.75 & +0.01 & \textbf{-0.93} & \textbf{-0.05} & \textbf{+1.15} & +0.01 \\
 & 4 & 1.18 & \textbf{-0.04} & \textbf{-0.78} & \textbf{-0.11} & \textbf{+0.28} & \textbf{-0.09} \\
\addlinespace[2pt]
Qwen2.5-7B & 2 & 1.53 & +0.01 & \textbf{-0.27} & \textbf{+0.06} & \textbf{+0.41} & \textbf{+0.04} \\
 & 3 & 1.44 & \textbf{-0.11} & \textbf{-0.29} & -0.02 & \textbf{+0.99} & \textbf{+0.20} \\
 & 4 & 0.88 & +0.00 & +0.03 & +0.00 & \textbf{+0.63} & \textbf{+0.14} \\
\addlinespace[2pt]
Qwen2.5-14B & 2 & 2.42 & \textbf{+0.05} & \textbf{-0.48} & \textbf{-0.04} & \textbf{+0.46} & \textbf{-0.10} \\
 & 3 & 1.60 & +0.01 & \textbf{-0.17} & +0.01 & \textbf{+1.78} & \textbf{-0.05} \\
 & 4 & 1.29 & \textbf{+0.13} & \textbf{-0.21} & \textbf{+0.07} & \textbf{+0.76} & \textbf{+0.18} \\
\addlinespace[2pt]
\bottomrule
\end{tabular}

\end{table}

\begin{table}[h]
  \centering\small
  \caption{Section~\ref{sec:causal} results not shown in Fig.~\ref{fig:causal}, at N = 2 and 3. Temperature: change in \dprime{} when the top three heads are flattened ($\tau=2$), and the temperature applied to every head that gives the highest \dprime{} (tested: 0.5, 0.67, 0.8, 1, 1.25, 1.5, 2). Suppression: false-alarm and hit rates before and after removing half of each distractor's letter-identity component. Target removed: change in \dprime{} when the target's own component is removed instead. Qwen2.5-14B was not run in the suppression sweep.}
  \label{tab:l5}
  \setlength{\tabcolsep}{4.5pt}
  \begin{tabular}{lrrrrrrrr}
\toprule
& & \multicolumn{2}{c}{temperature} & \multicolumn{4}{c}{suppress distractor content (half)} & target \\
\cmidrule(lr){3-4}\cmidrule(lr){5-8}
& & top 3, $\tau{=}2$ & all heads & \multicolumn{2}{c}{false alarms} & \multicolumn{2}{c}{hits} & removed \\
Model & $N$ & $\Delta d'$ & best $\tau$ & before & after & before & after & $\Delta d'$ \\
\midrule
Pythia-70m & 2 & -1.04 & 1 & 0.27 & 0.18 & 0.82 & 0.84 & -1.49 \\
 & 3 & -1.73 & 1 & 0.28 & 0.13 & 0.79 & 0.63 & -1.62 \\
Pythia-160m & 2 & -0.85 & 1.25 & 0.18 & 0.05 & 0.82 & 0.92 & -1.44 \\
 & 3 & -0.92 & 1.25 & 0.25 & 0.10 & 0.89 & 0.82 & -2.19 \\
Pythia-410m & 2 & -0.36 & 1.25 & 0.18 & 0.07 & 0.80 & 0.81 & -1.36 \\
 & 3 & -0.30 & 1.25 & 0.25 & 0.08 & 0.72 & 0.67 & -1.15 \\
Pythia-1b & 2 & -0.94 & 1.25 & 0.21 & 0.07 & 0.90 & 0.88 & -1.67 \\
 & 3 & -0.78 & 1.25 & 0.21 & 0.10 & 0.73 & 0.67 & -1.30 \\
Pythia-1.4b & 2 & -0.58 & 1.25 & 0.15 & 0.03 & 0.88 & 0.90 & -1.87 \\
 & 3 & -0.20 & 1.25 & 0.24 & 0.12 & 0.76 & 0.74 & -1.40 \\
Pythia-2.8b & 2 & -0.32 & 1.25 & 0.20 & 0.10 & 0.90 & 0.85 & -1.95 \\
 & 3 & -0.67 & 1 & 0.23 & 0.12 & 0.73 & 0.72 & -1.12 \\
Pythia-6.9b & 2 & -0.28 & 1.25 & 0.19 & 0.05 & 0.87 & 0.88 & -1.59 \\
 & 3 & -0.16 & 1.25 & 0.29 & 0.07 & 0.80 & 0.89 & -1.13 \\
GPT-2 & 2 & -0.77 & 0.8 & 0.28 & 0.08 & 0.93 & 0.77 & -2.11 \\
 & 3 & -0.42 & 1.25 & 0.30 & 0.13 & 0.77 & 0.81 & -1.45 \\
GPT-2 medium & 2 & -0.45 & 1.25 & 0.25 & 0.06 & 0.90 & 0.92 & -1.90 \\
 & 3 & -0.53 & 1 & 0.31 & 0.15 & 0.80 & 0.88 & -1.61 \\
GPT-2 large & 2 & -0.29 & 1 & 0.24 & 0.06 & 0.94 & 0.91 & -2.17 \\
 & 3 & -0.40 & 1 & 0.32 & 0.12 & 0.83 & 0.85 & -1.63 \\
GPT-2 xl & 2 & -0.20 & 1.5 & 0.19 & 0.10 & 0.73 & 0.76 & -1.42 \\
 & 3 & -0.32 & 1.25 & 0.21 & 0.06 & 0.55 & 0.57 & -1.04 \\
Qwen2.5-0.5B & 2 & -0.05 & 1.25 & 0.22 & 0.10 & 0.81 & 0.80 & -1.66 \\
 & 3 & -0.03 & 1 & 0.28 & 0.09 & 0.68 & 0.65 & -1.00 \\
Qwen2.5-1.5B & 2 & -0.11 & 1.25 & 0.15 & 0.04 & 0.88 & 0.91 & -2.04 \\
 & 3 & -0.12 & 0.8 & 0.22 & 0.08 & 0.83 & 0.93 & -1.63 \\
Qwen2.5-7B & 2 & -0.08 & 1.5 & 0.19 & 0.16 & 0.74 & 0.83 & -1.12 \\
 & 3 & -0.06 & 1 & 0.23 & 0.09 & 0.74 & 0.87 & -1.14 \\
Qwen2.5-14B & 2 & -0.17 & 1 & -- & -- & -- & -- & -- \\
 & 3 & -0.05 & 1.5 & -- & -- & -- & -- & -- \\
\midrule
median & & -0.32 & -- & 0.23 & 0.09 & 0.81 & 0.83 & -1.54 \\
\bottomrule
\end{tabular}

\end{table}

\begin{table}[h]
  \centering\small
  \caption{Locating analysis (share of misses with attention off target, on target but lost in later layers, or on target with the heads' output already wrong) and probe results (AUC of the model's own preference and of the probe for matches versus lures, and share of the model's lure false alarms the probe classifies as lures) per model.}
  \label{tab:l7l8}
  \setlength{\tabcolsep}{4.5pt}
  \begin{tabular}{lrrrrrrr}
\toprule
& & \multicolumn{3}{c}{share of misses} & \multicolumn{3}{c}{match vs.\ lure} \\
\cmidrule(lr){3-5}\cmidrule(lr){6-8}
Model & $N$ & elsewhere & lost later & head wrong & model AUC & probe AUC & FA called lure \\
\midrule
Pythia-70m & 2 & 0.08 & 0.49 & 0.43 & 0.58 & 0.97 & 0.75 \\
 & 3 & 0.01 & 0.42 & 0.57 & 0.57 & 0.91 & 0.61 \\
Pythia-160m & 2 & 0.26 & 0.34 & 0.40 & 0.63 & 0.98 & 0.82 \\
 & 3 & 0.03 & 0.59 & 0.38 & 0.58 & 0.99 & 0.85 \\
Pythia-410m & 2 & 0.11 & 0.36 & 0.53 & 0.66 & 1.00 & 0.95 \\
 & 3 & 0.06 & 0.40 & 0.54 & 0.58 & 0.98 & 0.84 \\
Pythia-1b & 2 & 0.11 & 0.32 & 0.57 & 0.68 & 1.00 & 0.95 \\
 & 3 & 0.06 & 0.30 & 0.64 & 0.60 & 0.98 & 0.87 \\
Pythia-1.4b & 2 & 0.12 & 0.72 & 0.16 & 0.76 & 1.00 & 0.99 \\
 & 3 & 0.06 & 0.38 & 0.56 & 0.62 & 0.99 & 0.92 \\
Pythia-2.8b & 2 & 0.04 & 0.36 & 0.60 & 0.71 & 1.00 & 1.00 \\
 & 3 & 0.04 & 0.38 & 0.58 & 0.60 & 1.00 & 0.96 \\
Pythia-6.9b & 2 & 0.19 & 0.55 & 0.26 & 0.74 & 1.00 & 1.00 \\
 & 3 & 0.08 & 0.38 & 0.54 & 0.60 & 1.00 & 0.98 \\
GPT-2 & 2 & 0.16 & 0.65 & 0.19 & 0.49 & 0.97 & 0.84 \\
 & 3 & 0.12 & 0.38 & 0.50 & 0.48 & 0.84 & 0.53 \\
GPT-2 medium & 2 & 0.18 & 0.53 & 0.29 & 0.48 & 0.99 & 0.87 \\
 & 3 & 0.07 & 0.39 & 0.54 & 0.47 & 0.86 & 0.57 \\
GPT-2 large & 2 & 0.72 & 0.15 & 0.13 & 0.56 & 0.99 & 0.92 \\
 & 3 & 0.54 & 0.20 & 0.26 & 0.50 & 0.88 & 0.59 \\
GPT-2 xl & 2 & 0.15 & 0.42 & 0.43 & 0.60 & 1.00 & 0.92 \\
 & 3 & 0.10 & 0.28 & 0.62 & 0.49 & 0.89 & 0.69 \\
Qwen2.5-0.5B & 2 & 0.16 & 0.50 & 0.34 & 0.62 & 1.00 & 1.00 \\
 & 3 & 0.09 & 0.23 & 0.67 & 0.53 & 0.99 & 0.91 \\
Qwen2.5-1.5B & 2 & 0.17 & 0.42 & 0.41 & 0.78 & 1.00 & 1.00 \\
 & 3 & 0.04 & 0.39 & 0.57 & 0.67 & 0.99 & 0.86 \\
Qwen2.5-7B & 2 & 0.14 & 0.43 & 0.43 & 0.63 & 1.00 & 0.98 \\
 & 3 & 0.45 & 0.19 & 0.36 & 0.60 & 0.98 & 0.88 \\
Qwen2.5-14B & 2 & 0.68 & 0.17 & 0.14 & 0.81 & 1.00 & 1.00 \\
 & 3 & 0.51 & 0.13 & 0.36 & 0.64 & 1.00 & 0.95 \\
\bottomrule
\end{tabular}

\end{table}

\textbf{Post-hoc attention edits in the minimal model.} On the 720 one-layer models (the original 300 models and 420 variants; Appendix~\ref{app:methods}), a fixed split of attention between $i$ and $i-N$ gives \dprime{}~$\approx0$ at every N; keeping only the model's own weight on $i$ and $i-N$ lowers \dprime{} from 2.70 to 0.63 at N = 1 and from 2.18 to 1.32 at N = 6, with false alarms rising from 0.03 to 0.67 at N = 1. Removing either the content or the position term of the scores gives \dprime{}~$\approx0$ (content removed: $|d'|\le0.04$; position removed: $-0.28$ to $-0.03$, because content alone cannot tell distance N from any other repeat). Sharpening the trained attention lowers \dprime{} at N = 1--5 and gains at most $+0.06$ at N = 6.

\end{document}